\documentclass[review]{elsarticle}
\usepackage{lineno,hyperref}
\modulolinenumbers[5]

\usepackage{times}
\usepackage{helvet}
\usepackage{courier}
\usepackage{url}
\usepackage{stfloats}
\usepackage{floatrow}
\usepackage{graphicx}
\usepackage{epstopdf}
\usepackage{algorithm}
\usepackage{algorithmic}
\usepackage{amsmath}
\usepackage{amssymb}
\usepackage{multirow}
\usepackage{wrapfig}
\usepackage{array}
\usepackage{subfigure}
\usepackage[usenames]{color}
\usepackage{wrapfig}
\usepackage{colortbl,booktabs}
\usepackage{verbatim}
\usepackage{rotating}
\usepackage{setspace}
\usepackage{caption}
\begin{document}
%

\title{Depth Quality-aware Selective Saliency Fusion for RGB-D Image Salient Object Detection}

\begin{frontmatter}



\title{Recursive Multi-model Complementary Deep Fusion for Robust Salient Object Detection via Parallel Sub Networks}


\author{Zhenyu Wu$^{1}$
        ~~~~Shuai Li$^{1}$
        ~~~~Chenglizhao Chen$^{1,2*}$
        ~~~~Aimin Hao$^1$
        ~~~~Hong Qin$^3$
        \\ $^1$Beihang University~~~~~~~~$^2$Qingdao University~~~~~~~~$^3$Stonybrook University}


\begin{abstract}
Fully convolutional networks have shown outstanding performance in the salient object detection (SOD) field.
The state-of-the-art (SOTA) methods have a tendency to become deeper and more complex, which easily homogenize their learned deep features, resulting in a clear performance bottleneck.
In sharp contrast to the conventional ``deeper'' schemes, this paper proposes a ``wider'' network architecture which consists of parallel sub networks with totally different network architectures.
In this way, those deep features obtained via these two sub networks will exhibit large diversity, which will have large potential to be able to complement with each other.
However, a large diversity may easily lead to the feature conflictions, thus we use the dense short-connections to enable a recursively interaction between the parallel sub networks, pursuing an optimal complementary status between multi-model deep features.
Finally, all these complementary multi-model deep features will be selectively fused to make high-performance salient object detections.
Extensive experiments on several famous benchmarks clearly demonstrate the superior performance, good generalization, and powerful learning ability of the proposed wider framework.
\end{abstract}




\end{frontmatter}




\section{Introduction}

The objective of salient object detection is to identify the most visually distinctive object in the given image~\cite{ZhengZZ18}. As a preprocessing tool, the salient object detection (SOD) has a wide range of practical applications including visual tracking~\cite{hong2015online}, localization~\cite{HePZ17}, video saliency~\cite{OurTIP19}, image captioning~\cite{xu2015show,fang2015captions}, image retrieval~\cite{gao20123}, visual question answering~\cite{lin2017task} and object retargeting~\cite{vinyals2015show}.


Previous works frequently treat the SOD as a multi-level perception task~\cite{itti1998model}, in which its key rationale is to make full use of the saliency clues at different perception levels~\cite{shen2012unified}.
Recently, the fully convolutional networks (FCNs) has been adopted for the robust SOD, in which such success should largely attributed to its ability to learn hierarchical saliency clues.
Thus, the current state-of-the-art (SOTA) models~\cite{hou2017deeply,wang2017stagewise,hu2018recurrently} generally focus on how to utilize the hierarchical deep features in ``single network'' to produce the high-quality SODs.
Nevertheless, the hierarchical deep features revealed in an identical network have a tendency to be homogenization, resulting in a limited performance eventually.

In the view of the neuroscience, the human visual system mainly comprises two largely independent subsystems that mediate different classes of visual behaviors~\cite{visualParallel,schiller1991parallel}.
The subcortical projection from the retina to cerebral cortex is strongly dominated by the two pathways that are relayed by the magnocellular (M) and parvocellular (P) subdivisions of the lateral geniculate nucleus (LGN), in which the Parallel pathways generally exhibit two main characteristics:
\underline{1)} the M cells contribute to the low-level transient processing (e.g., visual motion perception, eye movement, etc.) while the P cells contribute more to the high-level recognition tasks (e.g., object recognition, face recognition, etc.);
\underline{2)} the M and P cells are separated in the LGN, but it is recombined in visual cortex latter.

\begin{figure*}[!t]
\centering
\includegraphics[width=1\textwidth]{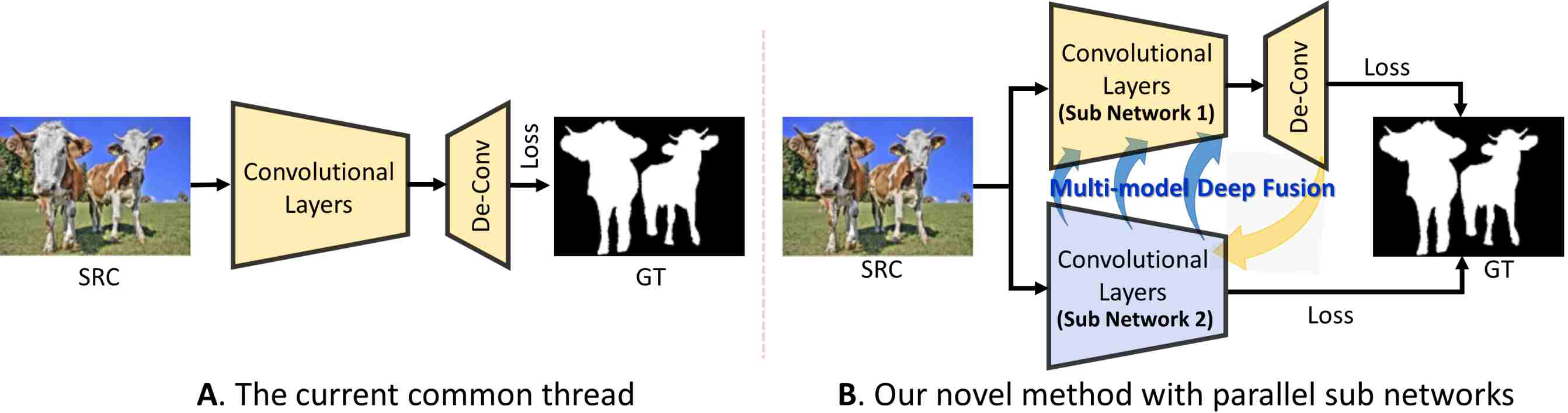}
\caption{\normalsize{The major difference between our method and the conventional methods}.}
\label{motivation}
\end{figure*}

Motivated by the above-mentioned theory, we propose to use two parallel networks (see Fig.~\ref{motivation}) to mimic the binocular vision of human visual system.
The key point of the proposed parallel network architecture is its ability to conduct multi-level saliency estimation while avoiding the conventional single network architecture inducted feature homogenization problem.
To achieve it, we devise a novel multi-model deep fusion framework, which attempts to fully exploit the complementary deep features from two different parallel sub networks; i.e., one for the coarse-level saliency localization and the other for the fine-scale detail polishing.
Meanwhile, inspired by the aforementioned attributes, we adopt the inter-model short-connections to recursively ensure a complementary status between each of our sub networks.
Moreover, we utilize a FCNs based saliency regressor to conduct selective deep fusion over those inter-model deep features, achieving a high-performance SOD eventually.

It should be noted that our ``wide'' scheme is solely implemented by using simple network architecture, yet it has achieved
remarkable performance improvement (e.g., averagely \underline{2}\% in F-max increasing and almost \underline{6}\% in MAE decreasing) comparing to the conventional complicated ``deeper'' schemes.
And such performance improvements are mainly induced by the newly designed multi-model fusion
scheme, in which the adopted simple network architecture is a hallmark of the proposed method. Moreover,
as far as we known, our paper is the first attempt to handle the SOD from the "wider" perspective.

To demonstrate the advantages of our method, we have conducted massive quantitative comparisons against 14 most representative SOTA methods over 5 widely used publicly available datasets.
Also, we have conducted extensive ablation studies to comprehensively verify the effectiveness of each essential component in our method.
Specifically, the salient contributions of this paper can be summarized as follows:

\begin{itemize}

\item We provide a deeper insight into the SOD task by imitating the binocular vision of human perception process;

\item To alleviate the obstinate feature homogenization problem in single network case, we utilize parallel sub networks to automatically reveal saliency clues at different spatial levels;

\item  We propose an end-to-end salient object detection model that learns diversity saliency clues in an iterative manner, aiming to achieve an optimal complementary status between the deep features extracted by our parallel sub networks;

\item We also provide a novel selective fusion strategy to fuse multi-model saliency clues for a high-performance salient object detection, archiving the new SOTA performance over the five adopted datasets.

\item The source code is available at: \textcolor{magenta}{https://github.com/Diamond101010/RMMDF}, which may has large potential to benefit the image salient object detection community in the future.
\end{itemize}


\section{Related Work}
\label{Related-Work}
\subsection{Deep Models with Single Network}
Early methods largely adopt various hand-crafted visual features~\cite{wei2012geodesic,li2013saliency,cheng2015global} to model the human visual attention~\cite{borji2015salient}.
After entering the deep learning era, deep learning based models \cite{cai2019saliency, xie2019high, wang2020salient, zhang2019salient, zhang2019hyperfusion, CC2015PR, CC2016PR} have been significantly improve saliency performance by using the automatically formulated multi-level deep features. 

Li \textit{et al}.~\cite{li2015visual} proposed a convolutional neural networks (CNNs) based computational model, which incorporates the multi-scale deep features via simply vector-wise feature concatenation.
Then, the same authors in \cite{li2017multi} further introduced a novel cascade network, which consists of several sub-networks to reveal saliency clues in a multi-level manner.
Similarly, Zhang~\textit{et al}.~\cite{zhang2017amulet} proposed a novel method to aggregate multi-level CNNs based deep features, in which the key rationale is to simultaneously integrate those high-level semantical information with those low-level details for the robust SOD.

Recently, the fully convolutional neural networks (FCNs) have achieved outstanding performance in many dense labeling tasks, including the SOD as well.
By using the FCNs based end-to-end saliency regression, both the efficiency and detection performance have been improved significantly.
Wang~\textit{et al.}~\cite{wang2016saliency} proposed to integrate hand-crafted features/priors into the recurrent fully convolutional networks (RFCNs).
Liu~\textit{et al.}~\cite{liu2016dhsnet} proposed a hierarchical refinement model to take full advantages of the coarse-level saliency clues to sharp the SOD boundaries.
Hou~\textit{et al.}~\cite{hou2017deeply} utilized the coarse-level deep features to facilitate the fine-level deep saliency computation via using the inter-layer short-connections.

\subsection{Deep Models with Parallel Sub Networks}
The recent development of network architecture has a tendency to become deeper and more complicate~\cite{TangW17}.
Zeiler \textit{et al.}~\cite{zeiler2014visualizing} have demonstrated that a deeper architecture can generally generate more discriminative features at the expense of more complexity architecture, leading the network difficult to train.
In sharp contrast to the ``deeper'' strategy, the ``wider'' architecture may become an intuitive choice, in this paper the term ``wider'' means to design network architecture with parallel sub networks.
For example, Lin \textit{et al.}~\cite{lin2015bilinear} proposed a bi-way architecture, utilizing two feature extractors to obtain multi-scale deep features for the image recognition.
Saito \textit{et al.}~\cite{saito2017dualnet} proposed a novel model for visual question answering, which attempts to learn discriminative features via using two independent sub networks to conduct feature extraction for multi-source data.
Kim \textit{et al.}~\cite{kim2018parallel} proposed to utilize a newly designed parallel feature pyramid network for object detection.

Recently, much effort has been made to design parallel architectures for the SOD.
Zhao \textit{et al.}~\cite{zhao2015saliency} designed a multi-context deep learning framework, in which the parallel revealed global context and local context are combined in an unified deep learning framework to jointly locate the salient object. Wang \textit{et al.}~\cite{wang2015deep} utilized parallel sub networks to respectively conduct pixel-level/object-level saliency computation, and then the revealed saliency clues will be fused as the SOD result.
Li \textit{et al.}~\cite{li2016deepsaliency} built a multi-task deep network to explore the common saliency consistency between the salient object detection and the semantic segmentation.



Actually, the ``wider'' choice has its merit to balance the trade-off between the saliency performance and the network complexity.
However, because the parallel structure adopted by the above mentioned methods are trained independently, those parallel learned deep features may not be able to effectively complement with each other, not to mention those occasional conflictions may lead the overall performance even worse.


In contrast to the above mentioned methods, the proposed model is completely different in 2 aspects:
\underline{1)} We utilize a novel recursive learning strategy to train parallel sub networks to obtain a complementary status between two subnetworks;
\underline{2)} As for those already learned complementary deep features, we utilize a selective fusion module to ensure an optimal fusion status for high-quality SOD result.

\section{Parallel Sub Networks}
\label{Method Overview}
Backbone networks with different structures may show
different fitting ability, and a good fitting ability frequently results in a strong semantical sensing ability. Thus, our
goal is to design a bi-stream network with two different
sub-networks, in which these sub-networks will potentially be able to provide complementary semantical deep features in terms of their different network structures and fitting abilities.

\begin{figure*}[!t]
\centering
\includegraphics[width=1\textwidth]{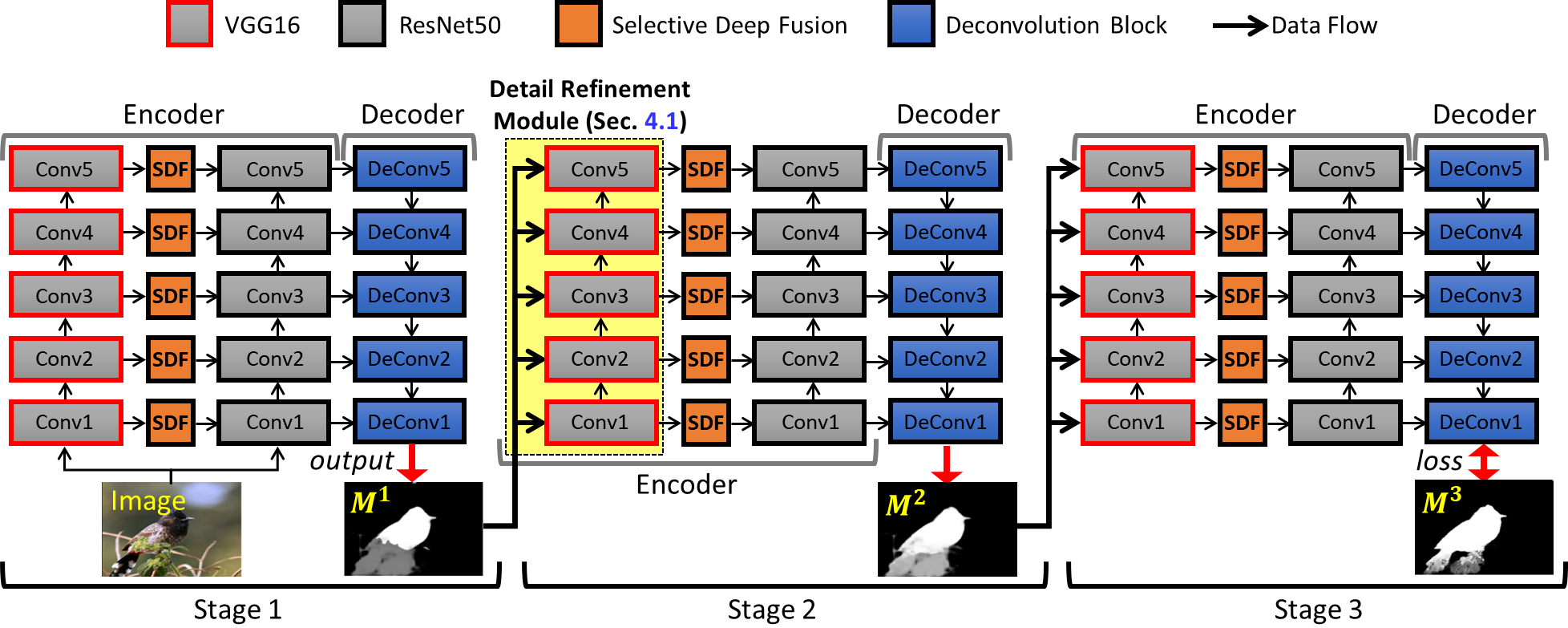}
\caption{The pipeline of our proposed method. Our network follows the encoder-decoder style, yet it different from previous methods, in which the encoder consists of two backbones with different structures, i.e., VGG16 and ResNet50. The input image is firstly passed through the Encoder to extract multi-scale convolutional deep features. Then, we use both the newly proposed Dense Aggregation Module (Sec.~\ref{sec:DAM}) and Selective Deep Fusion Module (Sec.~\ref{integrate module}) to make full use the multi-scale deep features which are extracted from VGG16 and ResNet50 respectively. The decoder network takes the multi-scale convolutional features as input to generate a finer saliency prediction $\textbf{M}^{t}$, which will latterly be refined by recursively using those low-level deep features in previous stage (Sec.~\ref{sec:DRM}). In each learning stage ($<$N), our method simultaneously uses the detail refinement module (to alleviate the spatial info loss)
and the dense aggregation module (to avoid the learning ambiguity) to ensure the complementary status between the parallel sub networks.
When our recursive learning reaches the final stage (=N), we
simultaneously feed the last feature layer of ResNet-50 and all side
layers of VGG-16 into the selective deep fusion network to produce the final SOD results.}
\label{pipeline}
\end{figure*}

Here we simply choose two vanilla networks, i.e., the ReNet-50 and the VGG-16, as our backbone sub networks, which can also be replaced by any other off-the-shelf networks.
Also, these two parallel sub networks will focus on different saliency perspective by using independent loss function to obtain diversity features.

As for the ResNet-50 sub network, we denote its convolutional blocks as: \{Conv1, Conv2, ..., Conv5\}, followed by five de-convolutional layers ($3\times3$).
Since each convolutional block will reduce the resolution of input feature map by 1/2, those feature maps produced by the  Conv5 block only have 1/32 resolution regarding the original input, which can merely locate the salient object, coarsely.
Thus, we utilize multiple de-convolutional layers to produce the fine-scale saliency score map $\textbf{M}^t$, where the superscript $t$ denotes the recursive learning stage.

The VGG-16 sub network is much simpler than the ResNet-50 sub network, and the VGG-16 only has five convolutional blocks: Conv1 (64 channels), Conv2 (128 channels), Conv3 (256 channels), Conv4 (512 channels) and Conv5 (512 channels). Each convolutional blocks is followed by a max-pooling layer of size 2 and a ReLU activation function.

Here we utilize $\textbf{X}$ = \{$\textbf{X}_i, i\in[1,5]$\} to denote the input maps for each convolutional block in the VGG-16 sub network, in which the $W_i$ and the $b_i$ respectively represent the predefined kernel and bias.
Thus, the learning procedure of our method can be uniformly formulated as Eq.~{{\ref{eq:X1}}}.
\begin{equation}
\label{eq:X1}
\textbf{X}_{i+1}\gets Conv(\textbf{X}_i):\ W_i^s*\textbf{X}_i + b_i,
\end{equation}
where $Conv(\cdot)$ denotes the convolutional operation and the superscript $s$ denotes the convolutional stride.
Similarly, we represent the input maps for convolutional blocks in our ResNet-50 sub network as $\textbf{F}$ = \{$\textbf{F}_i, i\in[1,5]$\}.

Fig.~{{\ref{pipeline}}} illustrates the overview of the proposed model, which mainly consists of three components: \underline{1)} detail refinement module; \underline{2)} dense aggregation module; and \underline{3)} selective deep fusion.
All these components will cooperate our recursive multi-model deep learning, which will be respectively introduced in the following sections.


\section{Inter-model Deep Fusion}
\label{method}
%
\subsection{Detail Refinement Module}
\label{sec:DRM}
Following the widely used encoder-decoder network architecture, the proposed detail refinement module (DRM) utilizes the ResNet-50 sub network to conduct an end-to-end saliency regression for the fine-scale saliency predictions, which will latterly be applied to another parallel sub network (VGG-16) to alleviate the spatial information loss problem, recursively.

\begin{figure*}[!t]
\centering
\includegraphics[width=1\textwidth]{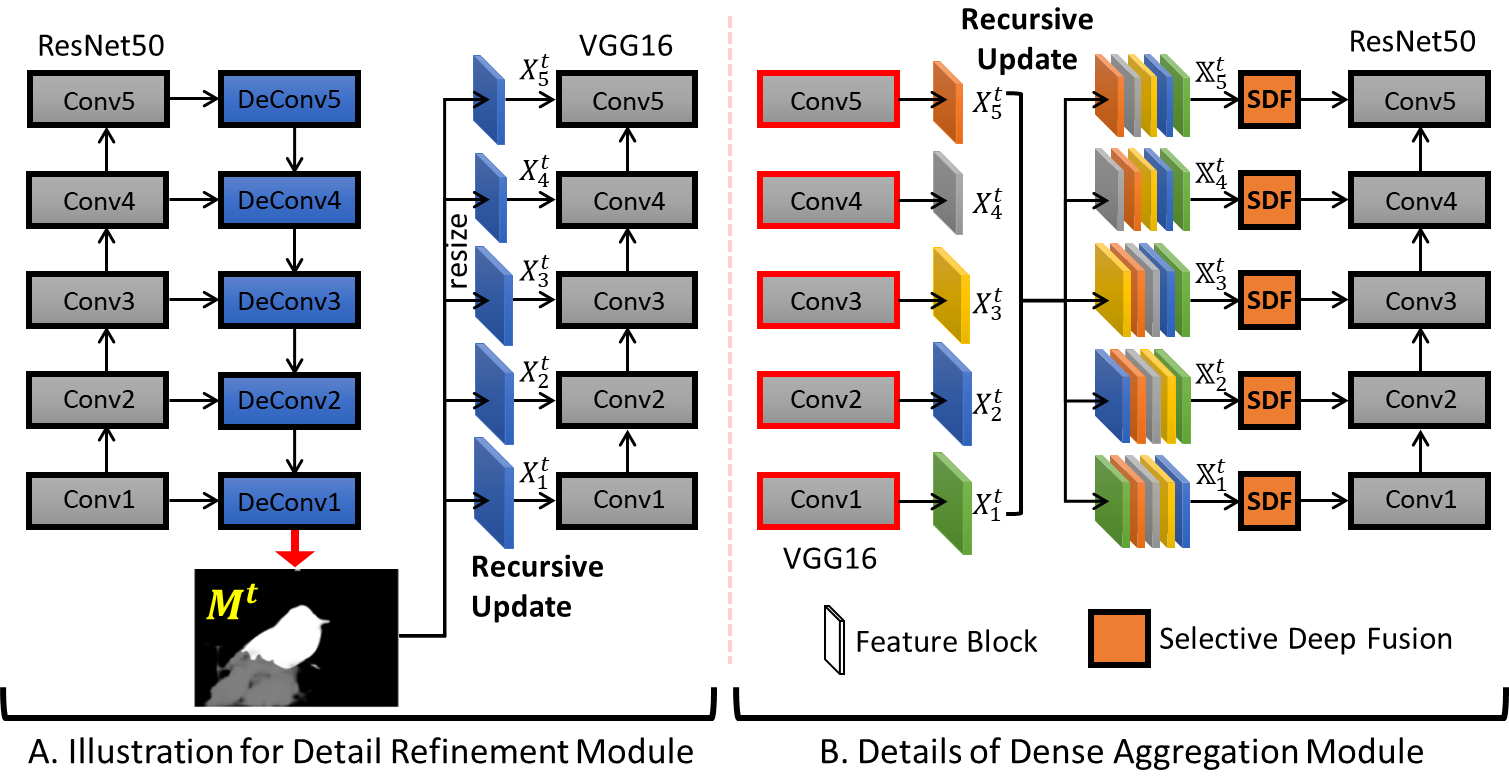}
\caption{The illustration of the proposed modules. The sub-figure A is the detailed architecture of the Detail Refinement Module (Sec.~\ref{sec:DRM}) in the $t$-th stage. We resize the $\textbf{M}^{t}$  to the same size of the $\textbf{X}^{t}_i$ and concatenate them together by performing convolutional operation. Then, the combined features will be feeded into the next stage, obtaining the $\textbf{M}^{t+1}$ with better details. The sub-figure B shows how to convert the multi-level deep features $\textbf{X}^{t}_i$ into the integrated feature maps $\mathbb{X}_i^t$, which will latterly prepare a set of finer deep features for the next learning stage (Sec.~\ref{sec:DAM}).}
\label{module}
\end{figure*}

Actually, the conventional networks usually adopt multiple convolution and pooling operations for their saliency regression, which easily degrade their performance due to the spatial information vanishes in deep layers. To alleviate it, Hou et al.~\cite{hou2017deeply} proposed to resort short-connections to integrate inter-layer deep features to compensate the lost spatial details.
However, deep features obtained by an identical single network have a tendency of homogenization, which heavily limits the complementary status between inter-layer deep features.

To further improve, we propose to construct dense connections between our parallel networks, see the pictorial demonstration in Fig.~\ref{module}-A.
Since the last layer of ResNet-50 can well represent the saliency details, we use it to recursively refine its parallel VGG-16 features ($\textbf{X}_i^t$, $i\in[1,2,3,4,5]$).
Also, we resize the resolution of $\textbf{M}^t$ according to each target block $\textbf{X}_i^t$, and then fuse these linked deep features by using a $3\times3$ convolution.
Here we formulate the recursively fusion procedure as Eq.~{{\ref{eq:FUpdating}}}.
\begin{equation}
\label{eq:FUpdating}
\textbf{X}_i^{t+1}\gets \left\{
\begin{array}{ll}
Conv\{\textbf{X}_i^{t}, \uparrow(\textbf{M}^t)\},\ if\ \xi(\textbf{M}^t)<\xi(\textbf{X}^t_i)\\
\\
Conv\{\textbf{X}_i^{t}, \downarrow(\textbf{M}^t)\},\ if\ \xi(\textbf{M}^t)>\xi(\textbf{X}^t_i)
\end{array} \right.,
\end{equation}
where $\uparrow(\cdot)$ and $\downarrow(\cdot)$ denote the up-sampling and down-sampling operations respectively, the function $\xi(\cdot)$ returns the feature size of the given input.

So far, by using Eq.~{{\ref{eq:FUpdating}}}, we have utilized the fine-scale saliency predicted by the ResNet-50 sub network to refine its parallel sub network VGG-16.
Meanwhile, in order to achieve an optimal inter-model complimentary status, those deep features of VGG-16 should also be used to shrink the problem domain of the ResNet-50 sub network.
Therefore, we recursively update \textbf{M} ($\textbf{M}^{t+1}\gets\textbf{M}^{t}$) in the ResNet-50 sub network.

\subsection{Dense Aggregation Module}
\label{sec:DAM}
Previous works~\cite{zhang2017amulet,hou2017deeply,hu2018recurrently} have shown that a good saliency model should take full advantage of its intermediate multi-level deep features, in which those high-level deep features usually concentrate on the high-level semantical information while those low-level features frequently focus on the subtle details.

Inspired by the aforementioned aspects, we attempt to utilize all those intermediate deep features in the VGG-16 sub network to recursively complement its parallel ResNet-50 network.
To this end, we propose a novel feature aggregation method named dense aggregation, see the pictorial demonstration in Fig.~{{\ref{module}}}-B.

For each recursive learning stage (i.e., noted by superscript $t$), we first utilize $1\times 1$ convolution to reduce the feature channel.
Thus, we can easily aggregate each feature block $\textbf{X}_i^t$ to 1 channel feature map $\hat{\textbf{X}}_i^t$.
Then, for each $\hat{\textbf{X}}_i^t$, we resize $\hat{\textbf{X}}_j^t$ ($j\ne i$) to be an identical size of $\hat{\textbf{X}}_i^t$ and aggregate all theses resized feature maps to an identical size of each ResNet-50' feature block $\textbf{F}_i^t$ by using $1\times 1$ convolution, which can be formulated as Eq.~{{\ref{eq:F}}}.
\begin{equation}\footnotesize
\mathbb{X}_i^t = \left\{
\begin{array}{ll}
Conv\{Cat(\hat{\textbf{X}}_1^t,\uparrow(\hat{\textbf{X}}_2^t),...,\uparrow(\hat{\textbf{X}}_5^t))\} & \textrm{\emph{if} $i=1$}\\
\\
Conv\{Cat(...,\downarrow(\hat{\textbf{X}}_{i-1}^t),\hat{\textbf{X}}_i^t,\uparrow(\hat{\textbf{X}}_{i+1}^t),...)\} & \textrm{\emph{if} $i=\{2,3,4\}$}\\
\\
Conv\{Cat(\downarrow(\hat{\textbf{X}}_{1}^t),...,\downarrow(\hat{\textbf{X}}_{4}^t),\hat{\textbf{X}}_5^t)\} & \textrm{\emph{if} $i=5$}\\
\end{array} \right.,
\label{eq:F}
\end{equation}
where $\uparrow(\cdot)$ and $\downarrow(\cdot)$ respectively denote the up-sampling/down-sampling operation, $Cat(\cdot)$ denotes the concatenation operation.

In general, those computed deep feature $\mathbb{X}_i^t$ ($i\in\{1,2,3,4,5\}$) can well represent the intermediate coarse-level saliency clues in the VGG-16 sub network, and we recursively aggregate these features into the ResNet-50 sub network as Eq.~{{\ref{eq:RUpdating}}}.
\begin{equation}
\textbf{F}_i^{t+1}\gets Conv(\textbf{F}_i^{t},\mathbb{X}_i^{t}),
\label{eq:RUpdating}
\end{equation}
where $\mathbb{X}_i^{t}$ denotes the resized $i$-th feature block in ResNet-50 at the $t$ learning stage.
Once the ResNet-50' deep features $\textbf{F}_i^{t}$ have been updated to $\textbf{F}_i^{t+1}$, we can achieve the improved fine-level SODs $\textbf{M}^{t+1}$ accordingly, which will be used to initiate another round of recursively learning in our detail refinement module.

In summary, there are totally three major advantages regarding the proposed dense aggregation module:\\
\underline{1)} Each coarse-level deep features generated from VGG-16 facilitate the computation of fine-scale saliency prediction of current ResNet-50 network, which ensures an effective complementary status between our parallel sub networks;\\
\underline{2)} The proposed dense aggregation scheme can correctly reveal the common consistency of those intermediate multi-level deep features, which making the fine-scale saliency prediction (ResNet-50) more accurate;\\
\underline{3)} The coarse-level deep features produced by VGG-16 can effectively shrink the problem domain of ResNet-50, boosting the convergency toward the true saliency.



\section{Selective Deep Fusion}
\label{integrate module}
For each step mentioned before, we assign a selective selective deep fusion module (SDF) to fuse those complementary deep features, which will be latter used to produce the high-quality SODs.


The conventional methods have well-investigated various hand-crafted fusion schemes (e.g., the multiplicative based ones, the additive based ones, and the maximum combination based ones) to combine saliency clues which are revealed at different spatial-levels.
However, these methods are elaborately designed for certain types of image scenes, which may fail to generalize well in other image scenes.
Therefore, we propose to utilize a newly designed selective deep fusion to handle the above-mentioned limitation.

As shown in Fig.~{{\ref{pipeline}}}, our selective deep fusion receives the deep features at the each recursive stage (we assign it to 3 according to the qualitative results demonstrated in Fig.~{{\ref{stage-wise}}} as inputs, i.e., $\textbf{M}^t$ , $\textbf{X}_i^t$, $i\in[1,5]$.



\begin{table*}
\centering
 \caption{{Details of our selective deep fusion module, in which the ``DeC.'' denotes the DeConv and the ``ConvC.'' denotes the Conv-Classifier.
 For simplicity, we have omitted the channel number of the ``Output'' because they have an identical channel number (i.e., 64), excepting for the last ConvC. which has 2 channels only}.}
\renewcommand\arraystretch{2}

 \label{tab:freq}
 \resizebox{\linewidth}{!}{
 \begin{tabular}{|c|c|c|c|c|c|c|c|c|c|}
   \toprule
   \Large{Layers} & \Large{Conv1} & \Large{Conv2} & \Large{Conv3} & \Large{Conv4} & \Large{DeC.4} & \Large{DeC.3} & \Large{DeC.2} & \Large{DeC.1} & \Large{ConvC.}\\
   \midrule
   \Large{Kernel} & \Large{33} &\Large{3$\times$3}& \Large{3$\times$3} &\Large{3$\times$3} & \Large{3$\times$3} &\Large{3$\times$3}&\Large{3$\times$3} &\Large{3$\times$3} & \Large{1$\times$1}\\
   \midrule
  	\Large{Channel} & \Large{64} &\Large{64} & \Large{64} &\Large{64} & \Large{64} &\Large{64}& \Large{64} &\Large{64} & \Large{2}\\
  \midrule
   \Large{Output} & \Large{256$\times$256} &\Large{128$\times$128} & \Large{64$\times$64} & \Large{32$\times$32} & \Large{32$\times$32} & \Large{64$\times$64} &\Large{128$\times$128}& \Large{256$\times$256} &\Large{256$\times$256} \\
 \bottomrule
	\end{tabular}}
\end{table*}

Since these deep features will gradually converge to the saliency ground truth as the recursive learning iteration goes on, it is intuitive to treat these deep features as individual saliency clues at the final recursive stage.
So, we combine the fine-scale saliency clue ($\textbf{M}^t$) with the convolved dense deep features Conv ($\mathbb{X}_i^t$) by using the element-wise summation in caffe, which can be formulated as Eq.~{{\ref{eq:sum}}}.
\begin{equation}
\textbf{S} = Sum\{Conv(\mathbb{X}_{1}^t),...,Conv(\mathbb{X}_{5}^t),\textbf{M}^t\},
\label{eq:sum}
\end{equation}
where \textbf{S} denotes the fused feature, which intrinsically contains complementary saliency clues of both parallel networks; the function $Sum\{\cdot\}$ denotes the element-wise operation to all its input, and each $\mathbb{X}_i^t$ is resized to an identical size of $\textbf{M}^t$ in advance.
Finally, we feed the combined feature \textbf{S} into a FCNs based saliency regression network, which consists of 4 convolutional blocks and 4 de-convolutional blocks to produce high-quality saliency maps.

We show the architecture details of the proposed selective deep fusion module in Tab.~{{\ref{tab:freq}}}.
Actually, this module is mainly consisted by two components: the encoder layers and the decoder layers. The encoder layers is composed of 13 convolutional layers.
Each of these convolution layers is followed by a batch normalization and a ReLU activation function.
Meanwhile, we assign each encoder layer with one corresponding decoder.
It also should be noted that we do not use any ReLU in the decoder layers.

Specially, our method totally uses three typical cross-entropy losses, i.e., two for the parallel sub networks and one for the selective fusion module, and we do not use any loss for our detail refinement module and dense aggregation module to avoid homogenizing deep features in our parallel networks.

\begin{figure*}[t]
\centering
\includegraphics[width=\linewidth]{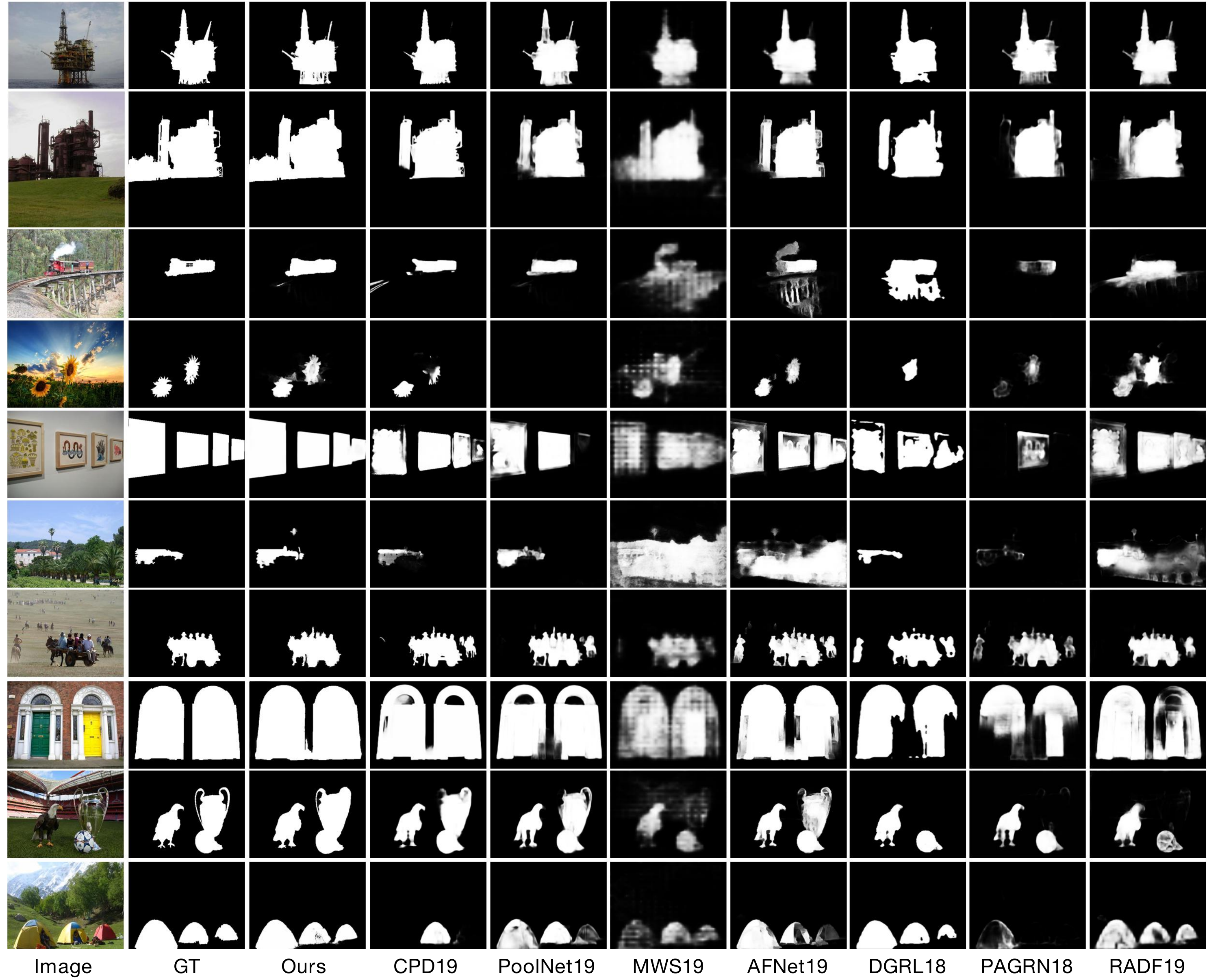}
\vspace{-0.4cm}
\caption{{Visual comparison of saliency maps. Note that GT stands for Ground truth. Apparently, It can be observed that the proposed model is able to handle diverse challenging scenes}.}
\label{saliencymap}
\end{figure*}

\begin{center}
\centering
\begin{table*}[!t]
\LARGE
\caption{Comparison of quantitative results including the max/average F-measure (larger is better) and MAE (smaller is better) on five well-known SOD benchmarks: DUT-OMRON~\cite{yang2013saliency}, DUTS-TE~\cite{zhao2015saliency}, ECSSD~\cite{yan2013hierarchical}, HKU-IS~\cite{zhao2015saliency} and PASCAL-S~\cite{li2014secrets}. The top three results are highlighted in {\textcolor{red}{red}}, {\textcolor{green}{green}}, and {\textcolor{blue}{blue}}, respectively.}
\renewcommand\arraystretch{1.7}

\resizebox{\linewidth}{!}{

\begin{tabular}{l|ccc|ccc|ccc|ccc|ccccc}

 \hline
\multirow{2}{*}{Method}
& \multicolumn{3}{c|}{DUT-OMRON}
& \multicolumn{3}{c|}{DUTS-TE}
& \multicolumn{3}{c|}{ECSSD}
& \multicolumn{3}{c|}{HKU-IS}
& \multicolumn{3}{c}{PASCAL-S}
\\
\cline{2-16}
&max$F_\beta$ & avg$F_\beta$ & MAE   &max$F_\beta$ &avg$F_\beta$ & MAE  &max$F_\beta$&avg$F_\beta$ & MAE &max$F_\beta$ &avg$F_\beta$ & MAE &max$F_\beta$ &avg$F_\beta$ & MAE\\
\hline
\textbf{Ours} & \textbf{\textcolor{red}{0.787}} &\textbf{\textcolor{red}{0.768}}&\textbf{\textcolor{red}{0.053}} & \textbf{\textcolor{blue}{0.839}} &\textbf{\textcolor{red}{0.798}}&  \textbf{\textcolor{blue}{0.045}} & \textbf{\textcolor{green}{0.925}}& \textbf{\textcolor{red}{0.905}} &\textbf{\textcolor{red}{0.042}} & \textbf{\textcolor{red}{0.918}}& \textbf{\textcolor{red}{0.911}} & \textbf{\textcolor{red}{0.031} } & \textbf{\textcolor{blue}{0.849}} & \textbf{\textcolor{red}{0.791}} &  \textbf{\textcolor{red}{0.089}} \\

CPD19~\cite{CPD}  & {0.754} &\textcolor{green}{0.738} &\textcolor{green}{0.056} & \textcolor{green}{0.841} & \textcolor{green}{0.784} &  \textcolor{green}{0.044} & {\textcolor{red}{0.926}} & \textcolor{blue}{0.880} &   \textcolor{green}{0.045} & \textcolor{blue}{0.911} & 0.883 & \textcolor{blue}{0.034} & {0.843} & {0.786} &  \textcolor{green}{0.092}\\

PoolNet19~\cite{PoolNet}& \textcolor{green}{0.763} & 0.683 & 0.071& \textcolor{red}{0.858}  &  \textcolor{blue}{0.781} & \textcolor{red}{0.040} & {0.920}&  \textcolor{blue}{0.880} & \textcolor{blue}{0.049}& \textcolor{green}{0.917}  &  \textcolor{green}{0.888} & \textcolor{green}{0.033} & \textcolor{red}{0.856} & \textcolor{green}{0.804} &   \textcolor{blue}{0.093}\\

AFNet19~\cite{AFNet}   & \textcolor{blue}{0.759}& \textcolor{blue}{0.729} &  \textcolor{blue}{0.057} & {0.838} & 0.772 &  0.046 &\textcolor{blue}{0.924}& 0.871 & \textcolor{red}{0.042}& {0.910} & 0.880 & 0.036 & \textcolor{green}{0.852} & 0.779 & \textcolor{red}{0.089} \\

MWS19~\cite{MWS}  & {0.677}& 0.606 & 0.109 & {0.722} & 0.686 &  0.092 & {0.859} & 0.838 &  0.096 & {0.835} & 0.813 &   0.084 & {0.781} & 0.743 & 0.153\\

PAGRN18~\cite{zhang2018progressive} & {0.707} &{0.709} & 0.071& {0.818}  &{0.782} & {0.056}& {0.904} & \textcolor{green}{0.887} &0.061 & {0.897} &\textcolor{blue}{0.885}& 0.048 &{0.817} & 0.721 & 0.120 \\

DGRL18~\cite{wang2018detect} & {0.739} &{0.723}& {0.062} & {0.806} &{0.769}& {0.051}& {0.914} & 0.866 &{0.049}& {0.900} & {0.882}& {0.036} & \textcolor{red}{0.856} &  0.785& {0.085}\\

RADF18~\cite{hu2018recurrently}   & {0.756}&{0.675} & {0.072} & {0.786} & {0.696} & {0.072} & {0.905} &  {0.861} &  0.060 & {0.895} & {0.848} &  {0.050} & {0.817} & \textcolor{blue}{0.790} & 0.123 \\

RFCN18~\cite{wang2018salient}  & {0.460}   & 0.466 &  0.138 & {0.478} & 0.479 & 0.136 & {0.656} & 0.649 & 0.161& {0.583} &  0.583 & 0.150 & {0.611}  & 0.596 &0.203\\

SRM17~\cite{wang2017stagewise}  &{0.725}   &{0.702} & {0.069} & {0.799} & {0.743} & {0.059} & {0.905} & {0.873} &  {0.054}& {0.893} & {0.870} &{0.046} & {0.812} & {0.786}& {0.105}  \\

Amulet17~\cite{zhang2017amulet} & {0.715}   &0.639 & 0.098 & {0.751} & 0.658& 0.085 & {0.904} & 0.849 &  {0.059} & {0.884} & 0.834 &0.052  & {0.836} & 0.744 & 0.107 \\

UCF17~\cite{zhang2017learning} & {0.705}& 0.602 & 0.132 & {0.740} & 0.611 & 0.118 & {0.897} &  0.819 &  0.078 & {0.871}& 0.800 & 0.074 & {0.820}  & 0.693 & 0.131 \\

DSS17~\cite{hou2017deeply} & {0.681} & 0.640 & 0.092 & {0.751} &  0.694 & 0.081 & {0.856} &  0.839 &  0.090 & {0.865} &  0.836 & 0.067 & {0.777} & 0.761& 0.149  \\

DeepSal16~\cite{li2016deepsaliency} & {0.749} & 0.640 &  0.084 & {0.749} &  0.637 & 0.090 & {0.892}&   0.816 & 0.079 & {0.849} & 0.786 & 0.080& {0.832} & 0.692 & 0.117 \\
MDF15~\cite{li2015visual}  & {0.642}   &  0.639 &  0.092 & {0.673} & 0.664 & 0.094 & {0.797} & 0.787 & 0.104 & {0.787} & 0.794 & 0.089 & {0.717} & 0.696  &0.172  \\

\hline
 \end{tabular}  }
\label{table1}
\end{table*}
\end{center}
\vspace{-1.5cm}
\section{Experiments and Results}
\label{experiments}
\subsection{Adopted Datasets}
We have evaluated our method on 5 widely used publicly available datasets, including DUT-OMRON~\cite{yang2013saliency} (with 5,168 images), DUTS-TE~\cite{zhao2015saliency} (with 5,019 images), ECSSD~\cite{yan2013hierarchical} (with 1,000 images), HKU-IS~\cite{zhao2015saliency} (with 4,447 images) and PASCAL-S~\cite{li2014secrets} (with 850 images).
Also, we have adopted 3 commonly used standard metrics to evaluate our method, including Precision-Recall (PR) curve, F-measure curve, and Mean Absolute Error (MAE).

%

\begin{figure*}[!t]
\centering
\includegraphics[width=\linewidth]{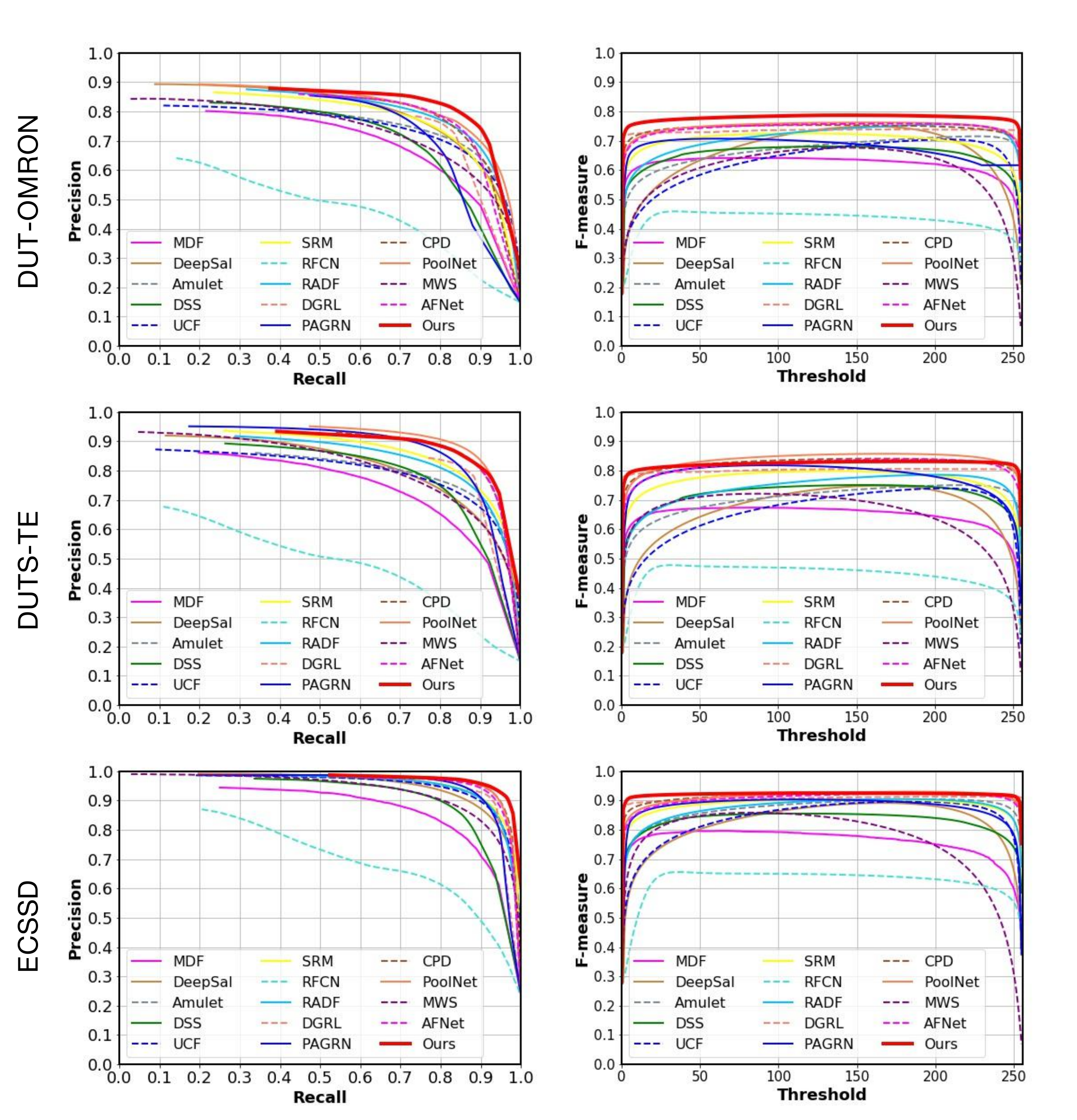}
\caption{Quantitative comparisons (PR curves and F-measure curves) between our method and 14 state-of-the-art methods over 5 adopted datasets, in which the left part is the PR curve and the right part is the F-measure curve.
Due to the limitation of space, we only provide the quantitative results over 3 datasets here, and the remaining 3 results can be found in Fig.~\ref{PRcurves2}.}
\label{PRcurves1}
\end{figure*}

\begin{figure*}[!t]
\centering
\includegraphics[width=\linewidth]{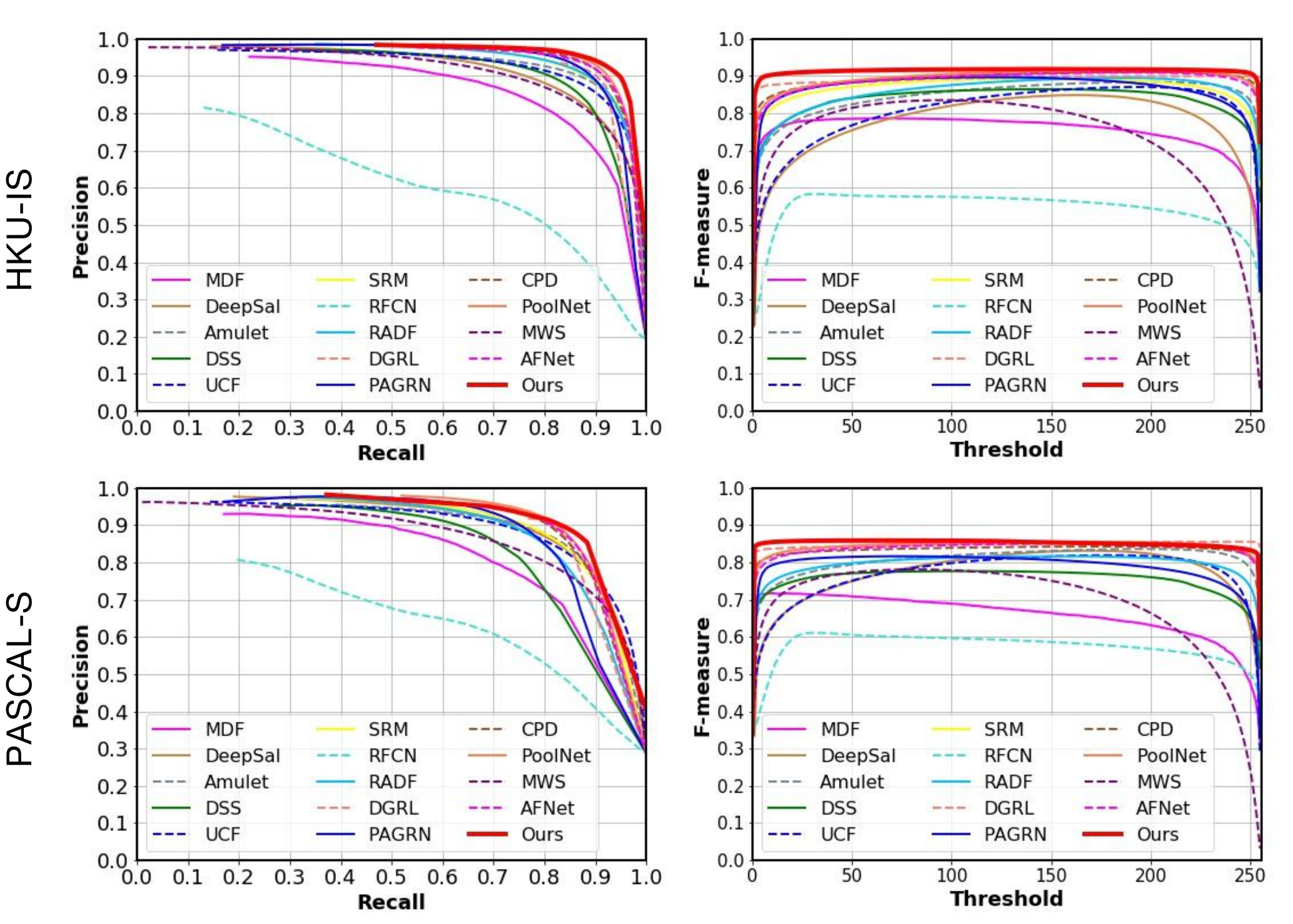}
\caption{Continued Quantitative comparisons (PR curves and F-measure curves) between our method and 14 state-of-the-art methods over 5 adopted datasets.}
\label{PRcurves2}
\end{figure*}

\subsection{Implementation Details}
The proposed method is developed on the public deep learning framework Caffe. We run our model in a quad-core PC with an i7-6700 CPU (3.4 GHz and 8 GB RAM ) and a NVIDIA GeForce GTX 1080 GPU (with 8G memory). Our model is trained on the MSRA10K dataset. Then, we test our model on other datasets. Due to the limited GPU memory, we set the mini-batch size to 4.
We use the stochastic gradient decent (SGD) method to train with a momentum 0.99, and the same weight decay 0.0005. Also, for our feature integration module, we use SGD with a momentum 0.9, and weight decay 0.0005. We set the learning rate as $10^{-8}$ and it reduces by a factor of 0.1 at 10k iterations. The training process of our model takes about 14 hours and converges after 5 epochs. During testing, the proposed model runs about 14 FPS with $256 \times 256$ resolution. The Tab.~{{\ref{time-analysis}}} shows the running time comparisons.

The performance improvements of our method are mainly brought
by the newly-designed multi-model fusion scheme, thus we can
implement the parallel sub networks using ``simple'' networks. For each sub network, the complexity/memory requirement
is better than the conventional single network cases, and the overall
complexity/memory requirement for our parallel sub networks is
comparable to the mainstream single network cases, e.g., by using a
GTX1080-8G GPU (with memory usage almost 100\%), it takes almost 14
hours to train our method, while the classic single method RADF18 takes about 10 hours.

\subsection{Quantitative Evaluation}
We have compared our method with 14 most representative SOTA methods, including MDF15~\cite{li2015visual}, DeepSal16~\cite{li2016deepsaliency},  Amulet17~\cite{zhang2017amulet}, DSS17~\cite{hou2017deeply},  UCF17~\cite{zhang2017learning}, SRM17~\cite{wang2017stagewise}, RFCN18~\cite{wang2018salient},RADF18~\cite{hu2018recurrently}, PAGRN18~\cite{zhang2018progressive}, DGRL18~\cite{wang2018detect}, MWS19~\cite{MWS}, CPD19~\cite{CPD}, AFNet19~\cite{AFNet}, and PoolNet19~\cite{PoolNet}. For all of these methods, we use the original codes with recommended settings or the saliency maps provided by the authors.

Both the qualitative comparisons and quantitative results can be found in Fig.~{{\ref{saliencymap}}} and Fig.~{{\ref{PRcurves1}}}.
As shown in Fig.~{{\ref{PRcurves1}}}, our method is competitive to the current SOTA methods across all datasets. As shown in Tab.~{{\ref{table1}}}, we utilize the averaged F-measure and the MAE to serve as the complementary evaluation metrics, which also indicate that our method consistently outperforms all other approaches.
As for the DUT-OMRON dataset, our model achieves $\underline{76.8}\%$ in F-measure and $\underline{5.3}\%$ in MAE while the second best (CPD19) only achieves $\underline{73.8}\%$ in F-measure and $\underline{5.6}\%$ in MAE.
Also, similar tendencies can be found in the HKU-IS dataset, which is one of the most challenge datasets.
Compared to the second best PoolNet19, our model increases $\underline{2.3}\%$ in F-measure and decreases $\underline{2}\%$ in MAE.

\begin{figure*}[!t]
\centering
\includegraphics[width=\linewidth]{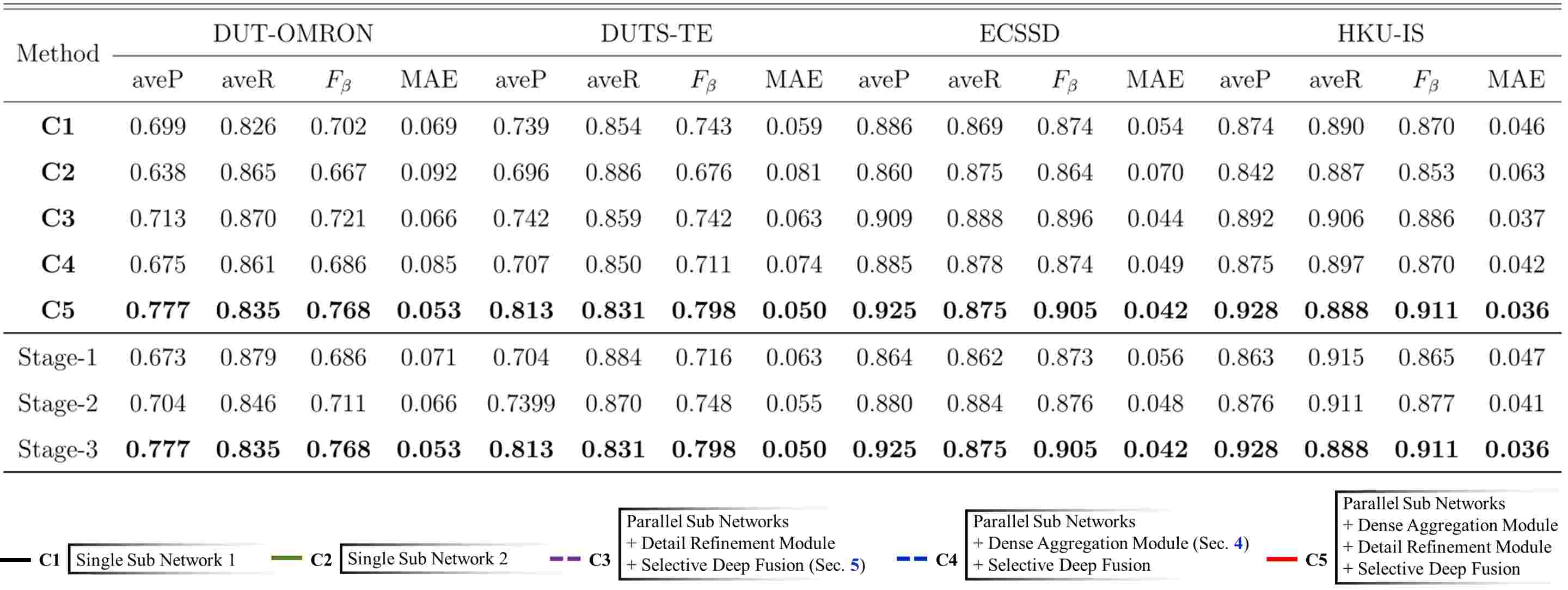}
\caption{Component evaluation results, including the average precision (aveP), average
recall (aveR), average F-measure scores and average MAE scores over DUT-OMRON, DUTS-TE, ECSSD
and HKU-IS dataset}
\label{ablation_pr}
\end{figure*}

\subsection{Component Evaluation}
To validate the effectiveness of our method, we have evaluated several variants in the proposed model by using different settings on the DUT-OMRON, DUTS-TE, ECSSD and HKU-IS dataset.
We start with two single-stream networks and progressively extend it with our newly designed modules, including the parallel backbones, the detail refinement module, the dense aggregation module and the selective deep fusion module.

As shown in in Fig.~{{\ref{ablation_pr}}}, our newly designed parallel architecture equipped with detail refinement module only (see the 3rd row) can achieve much better performance than the single sub network (the first row and 2nd row).
Meanwhile, the overall performance of the proposed parallel architecture with dense aggregation module can get the overall performance further improved, see the 4th row in Fig.~{{\ref{ablation_pr}}}. Specially, we can notice a significant performance improvement by introducing the proposed selective deep fusion module, see the 5th row.
All these results have demonstrated the effectiveness of the proposed method.

\begin{figure*}[!t]
\centering
\includegraphics[width=1\textwidth]{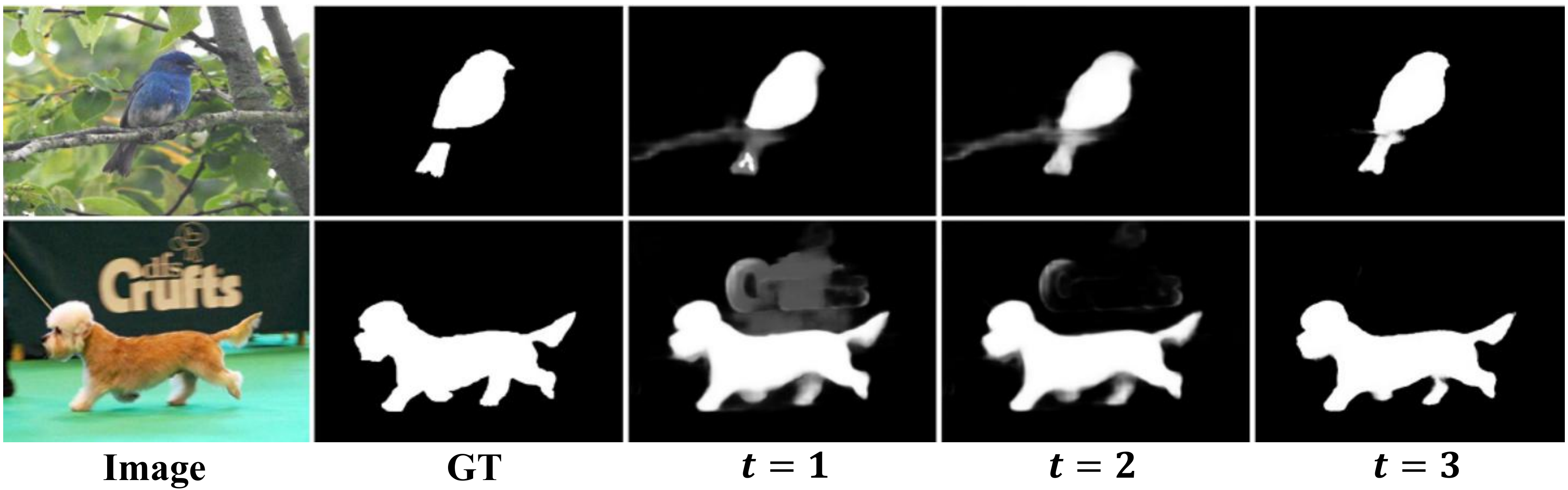}
\caption{Qualitative illustration of our recursive learning scheme, where $t$ denotes the saliency maps obtained via different learning stages.}
\label{stage-wise}
\end{figure*}


\subsection{Recursive Learning Validation}
As described in Sec.~{{\ref{method}}}, our method is trained in a recursive manner.
To validate the effectiveness of our stage-wise recursive learning scheme, we perform a detailed comparison of the proposed model at different recursive learning stages using max F-measure, average F-measure and MAE scores.
As shown in the last three rows of Fig.~{\ref{ablation_pr}}, the overall performance of our method becomes better as the stage-wise recursive learning continue, in which the corresponding qualitative demonstrations can be found in Fig.~{{\ref{stage-wise}}}.

\begin{table*}[!h]
\LARGE
\setlength{\tabcolsep}{2.5mm}
  \vspace{0.6cm}
  \caption{Runtime comparison (GPU time) with previous deep learning based saliency models.}

  \label{time-analysis}
  \resizebox{\textwidth}{!}{
  \begin{tabular}{ccccccccccccccl}
    \toprule
    Method   &Our   & 	CPD19 & AFNet19 &    DGRL18 & RADF18&
    		SRM17 & Amulet17 & UCF17 & DSS17 &
    		RFCN18& DeepSal16 & MDF15\\
    \midrule
   Time(s)   & 0.073  & 0.063 & 0.062 & 0.150 & 0.154 &
   			 0.070 &0.093 & 0.134 & 0.201 &
   			  4.72 & 0.150 &  19.278  \\
   	
  \bottomrule
  \label{table:time}
\end{tabular}}
\end{table*}

\subsection{Method Limitations}
Compared with previous works, our method can capture more powerful saliency clues from different saliency perspective while avoiding the obstinate feature conflictions by using the proposed multi-model fusion scheme.
As for images with clutter background, our method can well suppress those non-salient regions and preserves subtle salient details, which is proved by the increased precision rate and F-measure score in Fig.~{{\ref{PRcurves1}}}.

Nonetheless, we have noticed a slight decreasing regarding the average recall rate, which is mainly induced by an unbalanced bias in our multi-model fusion when computing those complementary deep features.
Another limitation of our model is the computational overhead for the stage-wise training. In the future, we plan to explore more efficient fusion approach by using the off-the-shelf model compression techniques to alleviate the computational burden.

\section{Conclusions}
\label{conclusion}
In this paper, we proposed a novel multi-model fusion method, in which two parallel sub networks are coordinated to learn complementary deep features recursively.
The key rationale of our method is to utilize two different sub networks to respectively concentrate on different saliency perspectives, while those revealed deep features shall be able to complementary with each other.
To achieve this goal, our method consists of three newly designed components: \underline{1)} Detail Refinement Module; \underline{2)} Dense Aggregation Module; and \underline{3)} Selective Deep Fusion Module.
We have utilized the detail refinement module to recursively compensate the lost spatial details, and then we have used the dense aggregation module to coarsely locate the salient object to shrink the given problem domain.
Meanwhile, we have adopted the short connections in our dense inter-model to ensure a complementary status between the parallel sub networks.

\newpage
\vspace{0.4cm}
\centerline{\textbf{\large References}}





\bibliographystyle{IEEEtran}
\bibliography{reference}

\begin{thebibliography}{10}
\providecommand{\url}[1]{#1}
\csname url@samestyle\endcsname
\providecommand{\newblock}{\relax}
\providecommand{\bibinfo}[2]{#2}
\providecommand{\BIBentrySTDinterwordspacing}{\spaceskip=0pt\relax}
\providecommand{\BIBentryALTinterwordstretchfactor}{4}
\providecommand{\BIBentryALTinterwordspacing}{\spaceskip=\fontdimen2\font plus
\BIBentryALTinterwordstretchfactor\fontdimen3\font minus
  \fontdimen4\font\relax}
\providecommand{\BIBforeignlanguage}[2]{{%
\expandafter\ifx\csname l@#1\endcsname\relax
\typeout{** WARNING: IEEEtran.bst: No hyphenation pattern has been}%
\typeout{** loaded for the language `#1'. Using the pattern for}%
\typeout{** the default language instead.}%
\else
\language=\csname l@#1\endcsname
\fi
#2}}
\providecommand{\BIBdecl}{\relax}
\BIBdecl

\bibitem{ZhengZZ18}
X.~Zheng, Z.~Zha, and L.~Zhuang, ``A feature-adaptive semi-supervised framework
  for co-saliency detection,'' in \emph{ACM International Conference on
  Multimedia}, 2018, pp. 959--966.

\bibitem{hong2015online}
S.~Hong, T.~You, S.~Kwak, and B.~Han, ``Online tracking by learning
  discriminative saliency map with convolutional neural network,'' in
  \emph{International Conference on Machine Learning}, 2015, pp. 597--606.

\bibitem{HePZ17}
X.~He, Y.~Peng, and J.~Zhao, ``Fine-grained discriminative localization via
  saliency-guided faster r-cnn,'' in \emph{ACM International Conference on
  Multimedia}, 2017, pp. 627--635.

\bibitem{OurTIP19}
C.~Chen, G.~Wang, C.~Peng, X.~Zhang, and H.~Qin, ``Improved robust video
  saliency detection based on long-term spatial-temporal information,''
  \emph{IEEE Transactions on Image Processing}, p.~1, 2019.

\bibitem{xu2015show}
K.~Xu, J.~Ba, R.~Kiros, K.~Cho, A.~Courville, R.~Salakhudinov, R.~Zemel, and
  Y.~Bengio, ``Show, attend and tell: Neural image caption generation with
  visual attention,'' in \emph{International Conference on Machine Learning},
  2015, pp. 2048--2057.

\bibitem{fang2015captions}
H.~Fang, S.~Gupta, F.~Iandola, R.~K. Srivastava, L.~Deng, P.~Doll{\'a}r,
  J.~Gao, X.~He, M.~Mitchell, J.~C. Platt \emph{et~al.}, ``From captions to
  visual concepts and back,'' in \emph{IEEE International Conference on
  Computer Vision and Pattern Recognition}, 2015, pp. 1473--1482.

\bibitem{gao20123}
Y.~Gao, M.~Wang, D.~Tao, R.~Ji, and Q.~Dai, ``3-d object retrieval and
  recognition with hypergraph analysis,'' \emph{IEEE Transactions on Image
  Processing}, vol.~21, no.~9, pp. 4290--4303, 2012.

\bibitem{lin2017task}
Y.~Lin, Z.~Pang, D.~Wang, and Y.~Zhuang, ``Task-driven visual saliency and
  attention-based visual question answering,'' \emph{arXiv preprint
  arXiv:1702.06700}, 2017.

\bibitem{vinyals2015show}
O.~Vinyals, A.~Toshev, S.~Bengio, and D.~Erhan, ``Show and tell: A neural image
  caption generator,'' in \emph{IEEE International Conference on Computer
  Vision and Pattern Recognition}, 2015, pp. 3156--3164.

\bibitem{itti1998model}
L.~Itti, C.~Koch, and E.~Niebur, ``A model of saliency-based visual attention
  for rapid scene analysis,'' \emph{IEEE Transactions on Pattern Analysis and
  Machine Intelligence}, vol.~20, no.~11, pp. 1254--1259, 1998.

\bibitem{shen2012unified}
X.~Shen and Y.~Wu, ``A unified approach to salient object detection via low
  rank matrix recovery,'' in \emph{IEEE International Conference on European
  Conference on Computer Vision}, 2012, pp. 853--860.

\bibitem{hou2017deeply}
Q.~Hou, M.-M. Cheng, X.~Hu, A.~Borji, Z.~Tu, and P.~Torr, ``Deeply supervised
  salient object detection with short connections,'' in \emph{IEEE
  International Conference on Computer Vision and Pattern Recognition}, 2017,
  pp. 5300--5309.

\bibitem{wang2017stagewise}
T.~Wang, A.~Borji, L.~Zhang, P.~Zhang, and H.~Lu, ``A stagewise refinement
  model for detecting salient objects in images,'' in \emph{IEEE International
  Conference on Computer Vision}, 2017, pp. 4019--4028.

\bibitem{hu2018recurrently}
X.~Hu, L.~Zhu, J.~Qin, C.~Fu, and P.~Heng, ``Recurrently aggregating deep
  features for salient object detection.'' in \emph{AAAI Conference on
  Artificial Intelligence}, 2018.

\bibitem{visualParallel}
W.~Merigan and J.~Maunsell, ``How parallel are the primate visual pathways?''
  \emph{Annual Review of Neuroscience}, vol.~16, no.~1, pp. 369--402, 1993.

\bibitem{schiller1991parallel}
P.~H. Schiller, N.~K. Logothetis, and E.~R. Charles, ``Parallel pathways in the
  visual system: their role in perception at isoluminance,''
  \emph{Neuropsychologia}, vol.~29, no.~6, pp. 433--441, 1991.

\bibitem{wei2012geodesic}
Y.~Wei, F.~Wen, W.~Zhu, and J.~Sun, ``Geodesic saliency using background
  priors,'' in \emph{IEEE International Conference on European Conference on
  Computer Vision}.\hskip 1em plus 0.5em minus 0.4em\relax Springer, 2012, pp.
  29--42.

\bibitem{li2013saliency}
X.~Li, H.~Lu, L.~Zhang, X.~Ruan, and M.-H. Yang, ``Saliency detection via dense
  and sparse reconstruction,'' in \emph{IEEE International Conference on
  Computer Vision}, 2013, pp. 2976--2983.

\bibitem{cheng2015global}
M.~Cheng, N.~J. Mitra, X.~Huang, P.~H. Torr, and S.~Hu, ``Global contrast based
  salient region detection,'' \emph{IEEE Transactions on Pattern Analysis and
  Machine Intelligence}, vol.~37, no.~3, pp. 569--582, 2015.

\bibitem{borji2015salient}
A.~Borji, M.~Cheng, H.~Jiang, and J.~Li, ``Salient object detection: A
  benchmark,'' \emph{IEEE Transactions on Image Processing}, vol.~24, no.~12,
  pp. 5706--5722, 2015.

\bibitem{cai2019saliency}
Q.~Cai, H.~Liu, Y.~Qian, S.~Zhou, X.~Duan, and Y.-H. Yang, ``Saliency-guided
  level set model for automatic object segmentation,'' \emph{Pattern
  Recognition}, vol.~93, pp. 147--163, 2019.

\bibitem{xie2019high}
W.~Xie, Y.~Shi, Y.~Li, X.~Jia, and J.~Lei, ``High-quality spectral-spatial
  reconstruction using saliency detection and deep feature enhancement,''
  \emph{Pattern Recognition}, vol.~88, pp. 139--152, 2019.

\bibitem{wang2020salient}
Q.~Wang, L.~Zhang, W.~Zou, and K.~Kpalma, ``Salient video object detection
  using a virtual border and guided filter,'' \emph{Pattern Recognition},
  vol.~97, 2020.

\bibitem{zhang2019salient}
Q.~Zhang, Z.~Huo, Y.~Liu, Y.~Pan, C.~Shan, and J.~Han, ``Salient object
  detection employing a local tree-structured low-rank representation and
  foreground consistency,'' \emph{Pattern Recognition}, vol.~92, pp. 119--134,
  2019.

\bibitem{zhang2019hyperfusion}
P.~Zhang, W.~Liu, Y.~Lei, and H.~Lu, ``Hyperfusion-net: Hyper-densely
  reflective feature fusion for salient object detection,'' \emph{Pattern
  Recognition}, vol.~93, pp. 521--533, 2019.

\bibitem{CC2015PR}
C.~Chen, S.~Li, H.~Qin, and A.~Hao, ``Real-time and robust object tracking in
  video via low-rank coherency analysis in feature space,'' \emph{Pattern
  Recognition}, vol.~48, pp. 2885--2905, 2015.

\bibitem{CC2016PR}
C.~Chen, S.~Li, A.~Hao, and H.~Qin, ``Robust salient motion detection in
  non-stationary videos via novel integrated strategies of spatio-temporal
  coherency clues and low-rank analysis,'' \emph{Pattern Recognition}, vol.~52,
  pp. 410--432, 2016.

\bibitem{li2015visual}
G.~Li and Y.~Yu, ``Visual saliency based on multiscale deep features,'' in
  \emph{IEEE International Conference on Computer Vision and Pattern
  Recognition}, 2015, pp. 5455--5463.

\bibitem{li2017multi}
X.~Li, F.~Yang, H.~Cheng, J.~Chen, Y.~Guo, and L.~Chen, ``Multi-scale cascade
  network for salient object detection,'' in \emph{ACM International Conference
  on Multimedia}, 2017, pp. 439--447.

\bibitem{zhang2017amulet}
P.~Zhang, D.~Wang, H.~Lu, H.~Wang, and X.~Ruan, ``Amulet: Aggregating
  multi-level convolutional features for salient object detection,'' in
  \emph{IEEE International Conference on Computer Vision}, 2017, pp. 202--211.

\bibitem{wang2016saliency}
L.~Wang, L.~Wang, H.~Lu, P.~Zhang, and X.~Ruan, ``Saliency detection with
  recurrent fully convolutional networks,'' in \emph{European Conference on
  Computer Vision}.\hskip 1em plus 0.5em minus 0.4em\relax Springer, 2016, pp.
  825--841.

\bibitem{liu2016dhsnet}
N.~Liu and J.~Han, ``Dhsnet: Deep hierarchical saliency network for salient
  object detection,'' in \emph{IEEE International Conference on Computer Vision
  and Pattern Recognition}, 2016, pp. 678--686.

\bibitem{TangW17}
Y.~Tang and X.~Wu, ``Salient object detection with chained multi-scale fully
  convolutional network,'' in \emph{ACM International Conference on
  Multimedia}, 2017, pp. 618--626.

\bibitem{zeiler2014visualizing}
M.~D. Zeiler and R.~Fergus, ``Visualizing and understanding convolutional
  networks,'' in \emph{IEEE European Conference on Computer Vision}.\hskip 1em
  plus 0.5em minus 0.4em\relax Springer, 2014, pp. 818--833.

\bibitem{lin2015bilinear}
T.-Y. Lin, A.~RoyChowdhury, and S.~Maji, ``Bilinear cnn models for fine-grained
  visual recognition,'' in \emph{IEEE International Conference on Computer
  Vision}, 2015, pp. 1449--1457.

\bibitem{saito2017dualnet}
K.~Saito, A.~Shin, Y.~Ushiku, and T.~Harada, ``Dualnet: Domain-invariant
  network for visual question answering,'' in \emph{IEEE International
  Conference on Multimedia and Expo}.\hskip 1em plus 0.5em minus 0.4em\relax
  IEEE, 2017, pp. 829--834.

\bibitem{kim2018parallel}
S.~Kim, H.~Kook, J.~Sun, M.~Kang, and S.~Ko, ``Parallel feature pyramid network
  for object detection,'' in \emph{European Conference on Computer Vision},
  2018, pp. 234--250.

\bibitem{zhao2015saliency}
R.~Zhao, W.~Ouyang, H.~Li, and X.~Wang, ``Saliency detection by multi-context
  deep learning,'' in \emph{IEEE International Conference on Computer Vision
  and Pattern Recognition}, 2015, pp. 1265--1274.

\bibitem{wang2015deep}
L.~Wang, H.~Lu, X.~Ruan, and M.~Yang, ``Deep networks for saliency detection
  via local estimation and global search,'' in \emph{IEEE International
  Conference on Computer Vision and Pattern Recognition}, 2015, pp. 3183--3192.

\bibitem{li2016deepsaliency}
X.~Li, L.~Zhao, L.~Wei, M.~Yang, F.~Wu, Y.~Zhuang, H.~Ling, and J.~Wang,
  ``Deepsaliency: Multi-task deep neural network model for salient object
  detection,'' \emph{IEEE Transactions on Image Processing}, vol.~25, no.~8,
  pp. 3919--3930, 2016.

\bibitem{yang2013saliency}
C.~Yang, L.~Zhang, H.~Lu, X.~Ruan, and M.-H. Yang, ``Saliency detection via
  graph-based manifold ranking,'' in \emph{IEEE International Conference on
  Computer Vision and Pattern Recognition}, 2013, pp. 3166--3173.

\bibitem{yan2013hierarchical}
Q.~Yan, L.~Xu, J.~Shi, and J.~Jia, ``Hierarchical saliency detection,'' in
  \emph{IEEE International Conference on Computer Vision and Pattern
  Recognition}, 2013, pp. 1155--1162.

\bibitem{li2014secrets}
Y.~Li, X.~Hou, C.~Koch, J.~M. Rehg, and A.~L. Yuille, ``The secrets of salient
  object segmentation,'' in \emph{IEEE International Conference on Computer
  Vision and Pattern Recognition}, 2014, pp. 280--287.

\bibitem{CPD}
Z.~Wu, L.~Su, and Q.~Huang, ``Cascaded partial decoder for fast and accurate
  salient object detection,'' in \emph{Proceedings of the IEEE Conference on
  Computer Vision and Pattern Recognition}, 2019, pp. 3907--3916.

\bibitem{PoolNet}
J.-J. Liu, Q.~Hou, M.-M. Cheng, J.~Feng, and J.~Jiang, ``A simple pooling-based
  design for real-time salient object detection,'' in \emph{Proceedings of the
  IEEE Conference on Computer Vision and Pattern Recognition}, 2019, pp.
  3917--3926.

\bibitem{AFNet}
M.~Feng, H.~Lu, and E.~Ding, ``Attentive feedback network for boundary-aware
  salient object detection,'' in \emph{Proceedings of the IEEE Conference on
  Computer Vision and Pattern Recognition}, 2019, pp. 1623--1632.

\bibitem{MWS}
Y.~Zeng, Y.~Zhuge, H.~Lu, L.~Zhang, M.~Qian, and Y.~Yu, ``Multi-source weak
  supervision for saliency detection,'' in \emph{Proceedings of the IEEE
  Conference on Computer Vision and Pattern Recognition}, 2019, pp. 6074--6083.

\bibitem{zhang2018progressive}
X.~Zhang, T.~Wang, J.~Qi, H.~Lu, and G.~Wang, ``Progressive attention guided
  recurrent network for salient object detection,'' in \emph{IEEE International
  Conference on Computer Vision and Pattern Recognition}, 2018, pp. 714--722.

\bibitem{wang2018detect}
T.~Wang, L.~Zhang, S.~Wang, H.~Lu, G.~Yang, X.~Ruan, and A.~Borji, ``Detect
  globally, refine locally: A novel approach to saliency detection,'' in
  \emph{IEEE International Conference on Computer Vision and Pattern
  Recognition}, 2018, pp. 3127--3135.

\bibitem{wang2018salient}
L.~Wang, L.~Wang, H.~Lu, P.~Zhang, and X.~Ruan, ``Salient object detection with
  recurrent fully convolutional networks,'' \emph{IEEE Transactions on Pattern
  Analysis and Machine Intelligence}, 2018.

\bibitem{zhang2017learning}
P.~Zhang, D.~Wang, H.~Lu, H.~Wang, and B.~Yin, ``Learning uncertain
  convolutional features for accurate saliency detection,'' in \emph{IEEE
  International Conference on Computer Vision}, 2017, pp. 212--221.

\end{thebibliography}

\end{document}